\documentclass[lettersize,journal]{IEEEtran}
\usepackage{xcolor}
\usepackage{amsmath,amsfonts}
\usepackage{algorithmic}
\usepackage{algorithm}
\usepackage{array}
\usepackage[caption=false,font=normalsize,labelfont=sf,textfont=sf]{subfig}
\usepackage[nolist]{acronym}
\usepackage{textcomp}
\usepackage{stfloats}
\usepackage{url}
\usepackage{verbatim}
\usepackage{graphicx}
\usepackage{cite}
\usepackage{standalone} %
\usepackage{tikz}
\usepackage{comment}
\usepackage{todonotes}
\usepackage{amsmath,amsfonts,bm}

\def\eqref#1{equation~\ref{#1}}

\def\1{\bm{1}}

\DeclareMathAlphabet{\mathsfit}{\encodingdefault}{\sfdefault}{m}{sl}
\SetMathAlphabet{\mathsfit}{bold}{\encodingdefault}{\sfdefault}{bx}{n}

\usepackage{hyperref}
\usepackage{booktabs}
\usepackage{multirow}
\usepackage{pifont}
\usepackage[numbers]{natbib}

\acrodef{dt}[DT]{Digital Twin}
\acrodef{aas}[AAS]{Asset Administration Shell}
\acrodef{bim}[BIM]{Building Information Modeling}
\acrodef{ssg}[SSG]{semantic scene graph}
\acrodef{uim}[UIM]{Unstructured Information Management}
\acrodef{vr}[VR]{virtual reality}
\acrodef{kg}[KG]{knowledge graph}
\acrodef{pel}[PEL]{Perceived Entity Linking}
\acrodef{mel}[MEL]{Multimodal Entity Linking}
\acrodef{ai}[AI]{artificial intelligence}
\acrodef{3dgs}[3DGS]{3D Gaussian Splatting}
\acrodef{nerf}[NeRF]{Neural Radiance Field}
\acrodef{pc}[PC]{Point Cloud}
\acrodef{icp}[ICP]{Iterative Closest Point}
\acrodef{slam}[SLAM]{Simultaneous Localization and Mapping}
\acrodef{sfm}[SfM]{Structure-from-Motion}
\acrodef{sgd}[SGD]{Stochastic Gradient Descent}
\acrodef{vlm}[VLM]{Vision-Language Model}
\acrodef{llm}[LLM]{Large Language Model}

\bstctlcite{IEEEexample:BSTcontrol}

\begin{document}    

\nocite{IEEEexample:BSTcontrol}

\title{Digital Twin Generation from Visual Data:\\A Survey}
\author{Andrew Melnik, Benjamin Alt, Giang Nguyen, Artur Wilkowski, Maciej Stefańczyk, Qirui Wu, Sinan Harms, Helge Rhodin, Manolis Savva, and Michael Beetz
\thanks{Andrew Melnik, Benjamin Alt, Giang Nguyen, and Michael Beetz are with the University of Bremen, Germany}
\thanks{Artur Wilkowski and Maciej Stefańczyk are with Warsaw University of Technology, Poland}
\thanks{Sinan Harms and Helge Rhodin are with Bielefeld University, Germany}
\thanks{Qirui Wu and Manolis Savva are with Simon Fraser University, Canada}
\thanks{Corresponding authors: Helge Rhodin (helge.rhodin@uni-bielefeld.de) and Andrew Melnik (andrew.melnik.papers@gmail.com)}
}

\maketitle
\newcommand{\fix}{\marginpar{FIX}}
\newcommand{\new}{\marginpar{NEW}}

\newcommand{\xmark}{\ding{55}} %
\newcommand{\cmark}{\ding{51}} %

\newcommand{\am}[1]{{\color{black}{#1}}}
\newcommand{\qw}[1]{{\color{black}{#1}}}
\newcommand{\ms}[1]{{\color{black}{#1}}}
\newcommand{\gn}[1]{{\color{black}{#1}}}
\newcommand{\sh}[1]{{\color{black}{#1}}}
\newcommand{\ba}[1]{{\color{black}{#1}}}
\newenvironment{aw}{}{}

\begin{abstract}
This survey examines recent advances in generating digital twins from visual data. These digital twins --- virtual 3D replicas of physical assets --- can be applied to robotics, media content creation, design or construction workflows. We analyze a range of approaches, including 3D Gaussian Splatting, generative inpainting, semantic segmentation, and foundation models, highlighting their respective advantages and limitations. In addition, we discuss key challenges such as occlusions, lighting variations, and scalability, as well as identify gaps, trends, and directions for future research. Overall, this survey aims to provide a comprehensive overview of state-of-the-art methodologies and their implications for real-world applications. Awesome Digital Twin: \url{https://awesomedigitaltwin.github.io}
\end{abstract}

\section{Introduction}
\label{introduction}

\IEEEPARstart{D}{igital} twins (\acsu{dt}s) refer to virtual representations of physical assets, that enable systems to simulate, monitor, and optimize real-world processes. The concept of a digital twin is applied across diverse domains, including manufacturing, construction and robotics. As definitions of digital twins vary widely between disciplines and applications, we adopt the general definition proposed by Vanderhorn et al. \cite{vanderhorn2021digital}: \textit{``a virtual representation of a physical system (...) that is updated through the exchange of information''}. In this survey, we focus on digital twins that replicate indoor environments using visual data. Traditionally, their creation has relied on specialized equipment such as LiDAR scanners or manual CAD modeling, which limit scalability and accessibility. Recent advances in computer vision and machine learning, notably Neural Radiance Fields (NeRFs) \cite{mildenhall2021nerf}, 3D Gaussian Splatting (3DGS) \cite{Kerbl2023}, and diffusion models \cite{melnik2024video}, have revolutionized this process, allowing the construction of detailed digital twins directly from video streams captured with consumer devices like smartphones. This emerging paradigm of video-based digital twinning opens new opportunities by bridging the real-to-sim and sim-to-real gaps between the physical and digital worlds.

\begin{figure}
    \centering
    \includegraphics[width=\linewidth]{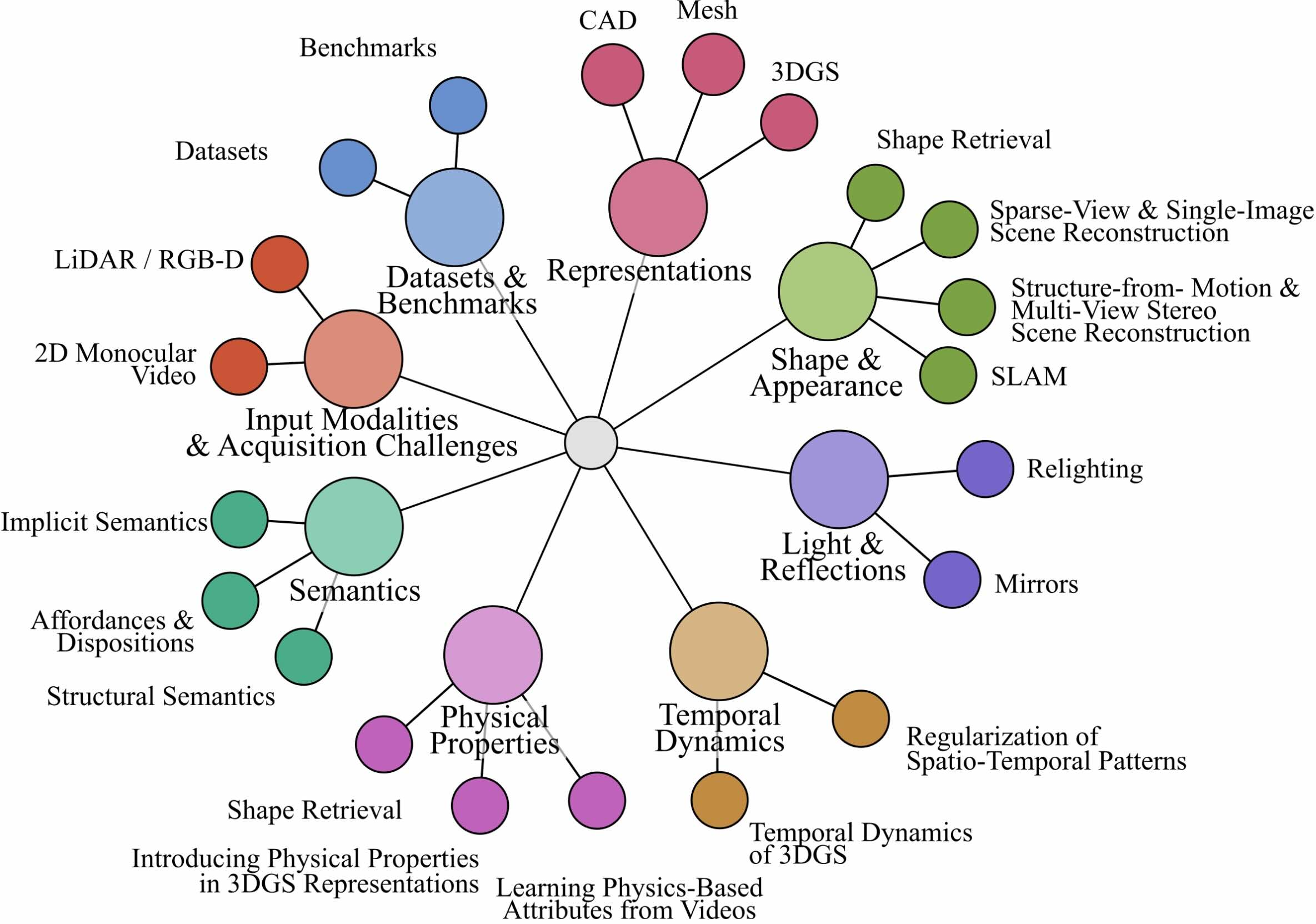}
    \caption{The structure of this paper, and a visual representation of the field.}
    \label{fig:digital-twins-taxonomy}
\end{figure}

Digital twins can range in accuracy from approximated representations (Digital Cousins \cite{dai2024automated}) to highly detailed virtual replicas of real-world environments. Recent advances in the automatic construction of digital twins from visual data provide opportunities for various domains, including robotics \cite{s2024splatrexperiencegoal, melnik2021using}, architecture \cite{abdelrahman2025digital}, industrial production \cite{sommer2023automated}, gaming \cite{Wang2024a}, video stream compression \cite{Zhang2024g}, and virtual reality \cite{Ye2024}.

The next frontier lies in enriching \ac{dt} models by integrating both the physical and semantic \cite{melnik2018world} properties of objects such as the articulation of parts, hierarchical structure and connectivity, interaction properties, rules, associated processes, material properties, audio patterns, etc.

This survey focuses on the generation of digital twins from images and videos, exploring the methods, challenges, and innovations that enable this process. Figure \ref{fig:digital-twins-taxonomy} provides a structure of this survey, which is oriented along the dimensions often modeled by digital twins. We aim to address key questions, including how to accurately reconstruct the geometry, appearance, dynamics, physics and semantics of indoor environments, and provide an overview over the available datasets and simulators to train and test digital twin construction pipelines. Our aim is to present approaches that not only enhance the fidelity of digital twins but also reduce the barriers to their adoption. We also highlight practical applications and outline future directions to advance the field further.

\section{Foundations of Digital Twin Representations}
\label{shape}

\am{
\subsection{Mesh \& CAD}

Explicit digital shape representations are usually based on mesh, CAD, \ac{pc} or \ac{3dgs} models - each having distinct advantages. \textbf{Mesh} representations use interconnected polygons (triangles or quads), making them efficient for real-time rendering, gaming, and 3D printing, though they lacks precise parametric control. \textbf{CAD} representations employ parametric curves (NURBS, B-Rep) to create highly accurate, editable, and feature-based models, making it ideal for engineering, manufacturing, and design applications.

\begin{table*}[ht!]
\caption{Comparison of \textbf{Mesh}, \textbf{CAD}, \textbf{NeRF} and \textbf{3DGS} models as representations for digital twins}
    \centering
    \small
    \begin{tabular}{|p{1.9cm}|p{4.7cm}|p{4cm}|p{4.9cm}|}
        \hline
        \textbf{Feature} & \textbf{3D Gaussian Splatting} & \textbf{Mesh Model} & \textbf{CAD Model} \\
        \hline
        \textbf{Representation} & Gaussian splats (point-based with density, color, and opacity) & Polygonal (triangles/quads) & Mathematical (NURBS, B-Rep) \\
        \hline
        \textbf{Precision} & Approximate (view-dependent, continuous representation) & Approximate & Highly precise \\
        \hline
        \textbf{Modifiability} & Hard to modify (requires re-optimization) & Hard to modify & Easy to modify (parametric) \\
        \hline
        \textbf{Usage} & Real-time rendering, neural scene representation, volumetric rendering & 3D graphics, gaming, animation, 3D printing & Engineering, manufacturing, design \\
        \hline
        \textbf{File Size} & Usually compact (but can grow with complexity) & Usually smaller & Larger due to parametric data \\
        \hline
        \textbf{Rendering} & Direct rendering via splatting (fast and efficient) & Rasterization or ray tracing & Mostly used for manufacturing, converted to mesh for visualization \\
        \hline
        \textbf{Conversion} & Difficult to convert to mesh or CAD & Can be converted to CAD (but loses precision) & Can be meshed or voxelized \\
        \hline
    \end{tabular}
    \label{table:comparison}
\end{table*}
}

\subsection{3D Gaussian Splatting}
\sh{
\acf{3dgs} \cite{Kerbl2023} is a radiance field-based explicit scene representation that models scenes using 3D Gaussian primitives with spatial and radiance properties, offering a more expressive alternative to traditional point clouds. Each 3D point in this representation is not just a simple coordinate, but a learnable 3D Gaussian distribution characterized by a covariance $\Sigma$ and mean $\mu$:
\begin{equation}
    G(x)=e^{-\frac{1}{2}(x-\mu)^{\top}\Sigma^{-1}(x-\mu)}
\end{equation}
The means of 3D Gaussians are initialized by a set of sparse point clouds (e.g. obtained from \ac{sfm}). Each Gaussian is parameterized by its center position $\mu \in \mathbb{R}^{3}$, spherical harmonics (SH) coefficient $c \in \mathbb{R}^{k}$ (with $k$ degrees of freedom) for view-dependent color, rotation factor $r \in \mathbb{R}^{4}$ (in quaternion rotation), scale factor $s \in \mathbb{R}^{3}$ and opacity $\alpha \in \mathbb{R}$. 
The covariance matrix $\Sigma$ describes an ellipsoid configured by a scaling matrix $S =\text{diag}([s_x, s_y, s_z])$ and rotation matrix $R=\text{q2R}([r_w, r_x, r_y, r_z])$ where q2R() is a function to construct a rotation matrix from a quaternion. The covariance matrix can then be computed as, 
\begin{equation}
    \Sigma = RSS^{\top}R^{\top}
\end{equation}
To enable differentiable optimization of 3D Gaussian parameters for accurate scene representation, the Gaussians are projected onto a 2D image plane via splatting \cite{Zwicker2002}. This involves applying a viewing transformation \( W \) and the Jacobian \( J \) of the affine approximation of the projection to the 3D covariance \( \Sigma \), yielding the 2D covariance \( \Sigma' = J W \Sigma W^{\top} J^{\top} \) in camera coordinates.
To achieve fast rendering and allow approximate $\alpha$-blending, the final image is split into $ 16 \times 16$ tiles (Figure \ref{fig:3dgs} c). Then, for each tile, the projected Gaussian that intersect with the tile's view frustum are sorted by depth (Figure \ref{fig:3dgs} d). Consequently, at a given pixel the color $C$ is computed by $\alpha$ compositing the opacity $\alpha$ and color $\mathbf{c}$, computed from spherical harmonics parameters, of overlapping 3D Gaussians (Figure \ref{fig:3dgs} e)
\begin{equation}
    C = \sum_{i \in \mathcal{N}} \mathbf{c}_i \alpha_i \prod_{j=1}^{i-1}(1-\alpha_j)
\end{equation}
During optimization, parameters are initialized from \ac{sfm} point clouds or random values. It employs \ac{sgd} \cite{Amari1993} with an L1 and D-SSIM \cite{Wang2004} loss function, optimizing against ground truth and rendered views. To address under- and over-reconstruction, periodic adaptive densification is applied, adjusting points with high gradient variations and eliminating those with low opacity (Figure \ref{fig:3dgs} f). This enhances scene representation quality while minimizing rendering errors.
\Ac{3dgs} excels in photorealistic, view-dependent rendering and is particularly useful for neural scene representations and volumetric graphics. Although mesh and CAD models are more structured and interpretable, \ac{3dgs} offers a continuous and memory-efficient alternative for immersive applications (see Table \ref{table:comparison}). The growing number of new methods for automated \ac{3dgs} reconstruction is establishing \ac{3dgs} as one of the most promising representations for approximating 3D scenes from visual input.
}

\subsection{Surfels}
\sh{
Surfels (surface elements) are point-based rendering primitives used to approximate surfaces through oriented point samples enriched with attributes such as position, normal vector, color or texture \cite{Pfister2000}. 
Recent efforts to overcome the limitations of 3D Gaussian Splatting have led to a renewed interest in surfel-based representations.

Despite their efficiency and photorealistic rendering, \ac{3dgs} shows limitations with surface reconstructions. 3D Gaussian splats are of volumetric, view-inconsistent nature, therefore limited in accurate geometric representations. To address this, recent methods share a unifying principle of transforming unstructured volumetric splats into compact, surface-aware, and view-consistent primitives such as disks, planes, or flattened ellipsoids (Figure \ref{fig:3dgs} g).
These methods introduce 2D-oriented Gaussians that lie explicitly on the scene's surfaces \cite{Huang2024a, Chen2024c} enforced by geometric regularization, e.g. depth-normal consistency or multi-view alignment. A parallel strategy, taking direct inspiration from surfels, collapses the z-scale of 3D Gaussians to create Gaussian surfels, enabling explicit normal encoding and enhanced optimization stability \cite{Dai2024}. Beyond geometric fidelity, other approaches emphasize compactness and rendering efficiency. By applying 2D Gaussians, ultra-fast image representation and compression can be achieved avoiding volumetric overhead while preserving detail \cite{Zhang2024g}. Other approaches, meanwhile, extend 2D Gaussian Splatting by attaching per-primitive textures, improving on high-frequency appearance modeling and disentangling geometry from texture \cite{Song2024a, Weiss2024}.
}

\section{Shape and Appearance}

\subsection{Shape Retrieval}

\begin{figure*}[t]
\centering
\includegraphics[width=0.9\linewidth]{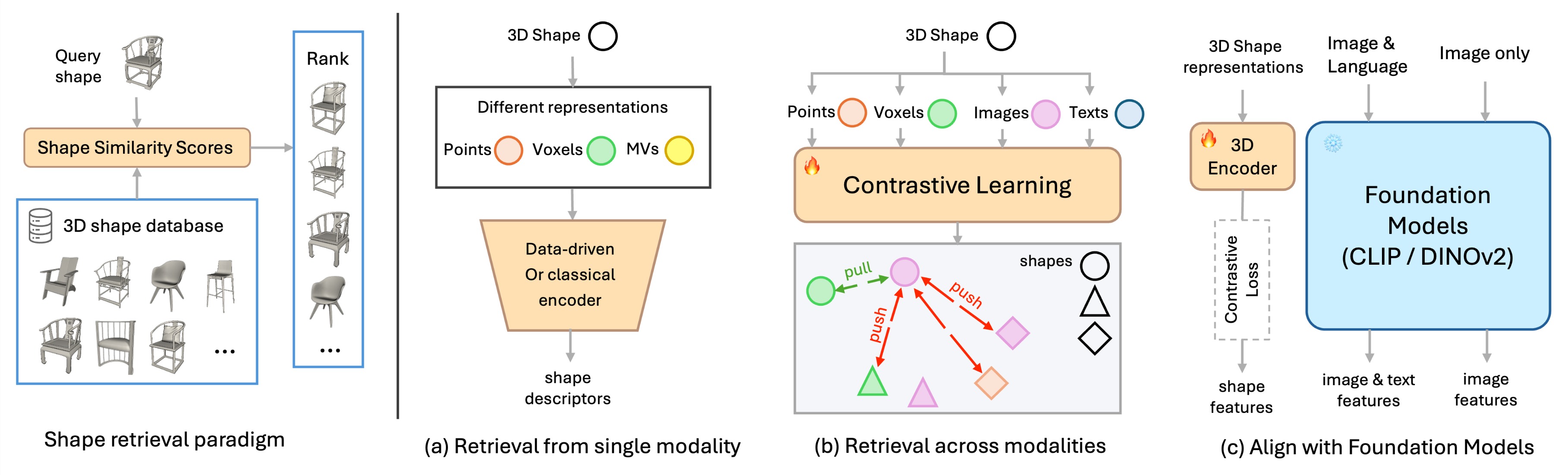}
\caption{Summary of the 3D shape retrieval paradigm [TODO: What paper is this from?]
} 
\label{fig:shape_retrieval}
\end{figure*}

\qw{
Typical 3D shape retrieval measurements require both the ground-truth shape and retrieved 3D shapes to compute either retrieval accuracy or point-based reconstruction scores, such as Chamfer Distance (CD), Normal Consistency (NC) and F-score $\mathrm{F}1^{t}$ conditioned on point-wise distance $t$ that indicates the strictness of points being accurately reconstructed. 
These metrics can be directly used as shape similarity to retrieve 3D shapes given an object point cloud as query.
We note that point-based metrics are sensitive to the point sampling method and the number of sampled points.
The work~\cite{wu2024generalizing} shows that Under the same sampling condition, NC and $\mathrm{F}1^{t}$ tend to rank unrelated shapes higher compared to CD, which either present distinguishable geometric structure or have wrong categories. Overall, CD is a more robust point cloud-based similarity score for shape retrieval that outputs reasonable shape ranking without the need to tweak hyperparameters.
Besides 3D representation with explicit geometry such as point clouds and voxel, 3D shapes can also be represented as multi-view images. 
A prescribed approach~\cite{chen2003visual} represent a 3D shape as light-field descriptor (LFD), computed from a set of pre-rendered binary masks, and retrieve the target shape by comparing L1 distance of features. 

More recently, learning-based methods are developed to advance the progress in different variants of the 3D shape retrieval task (see Figure \ref{fig:shape_retrieval}), involving querying of shapes from single-view images~\cite{
kuo2021patch2cad, lin2021single, rothganger2023shape}, text~\cite{
ruan2024tricolo}, voxels~\cite{gumeli2022roca}, point clouds~\cite{langer2024fastcad} and other 3D objects~\cite{he2018triplet}.
To align the different modalities of the same physical entity, queries and 3D objects are usually encoded into a joint-embedding space that learnt in an end-to-end manner using triplet or contrastive losses.
In such case, retrieval can be achieved by fetching the 3D object with the minimal distance in the high-dimensional embedding space to the query object. The distance is usually messured by L2 norm or cosine distance.
Given an image query of the target object, CMIC~\cite{lin2021single} achieves state-of-the-art performance on retrieval accuracy by using instance-level and category-level contrastive losses.
To handle different texture and lighting conditions in images, \cite{fu2020hard} proposed to synthesize textures on 3D shapes to generate hard negatives for additional training.
\cite{uy2021joint} used a deformation-aware embedding space so that retrieved models better match the target after an appropriate deformation.

The natural limitation of learning a image-CAD joint-embedding space is that it would consume large-scale image-shape pair annotations to achieve generalizability to novel object instances and categories, which is empirically infeasible and expensive. One promising solution is to investigate the use of \ac{vlm} (CLIP~\cite{radford2021learning}) to guide the alignment of 3D shape embeddings.
ULIP~\cite{xue2022ulip} is one of the pioneers that align text, image and 3D modalities. It uses frozen CLIP text-image encoders, and trained only the point-cloud encoder. However, the generalizability of these works were limited by the available dataset size at the time. OpenShape~\cite{liu2023openshape} and ULIP-2~\cite{xue2023ulip2} are among the first to scale up paired text-image-3D training data to train a point cloud encoder aligned with the CLIP embedding space multi-modality contrastive learning. They take advantage of combined large-scale 3D shape datasets~\cite{chang2015shapenet, objaverse} to recover real-world data distribution beyond a fixed number of classes and pretrained LVLMs (GPT-4v) for rendering captioning to generate paired text description.
Uni3D~\cite{zhou2023uni3d} maps point cloud patches to image patches within pre-trained ViTs to align the 3D point cloud features with the image-text aligned features. In contrast to aforementioned existing works, DuoDuoCLIP~\cite{lee2024duoduo} learns shape embeddings from multi-view images instead of point clouds that show better generalization performance and more light-weight hardware requirements.

}

\subsection{Structure-from-Motion \& Multi-View Stereo Scene Reconstruction}
\sh{
COLMAP \cite{schonberger2016structure} is a general-purpose \ac{sfm} pipeline. \ac{sfm}
reconstructs 3D scene geometry from 2D images taken from different viewpoints (Figure \ref{fig:3dgs}b). It detects and matches features across images, estimates camera poses and intrinsics, and triangulates 3D points \cite{schonberger2016structure}. The core task is to jointly recover scene geometry and camera motion. Feature detection involves identifying keypoints such as edges, corners, and textures, using methods such as SIFT
or more recent neural and transformer-based models like SuperGlue \cite{Sarlin2020}, R2D2 \cite{Revaud2019}, and LoFTR \cite{Sun2021}. Matched features across views are used to estimate camera parameters, enabling triangulation of 3D points. Bundle adjustment \cite{Triggs2000} refines both structure and motion by minimizing reprojection error. The output is a sparse 3D point cloud, a set of $(X,Y,Z)$ points representing surface geometry, optionally enriched with attributes such as color, intensity, or normals. These point clouds serve as input for further modeling tasks such as meshing or scene reconstruction. \ac{sfm} is often combined with Multi-View Stereo to produce dense point clouds with higher geometric detail. %

\begin{figure*}[ht!]
    \centering
    \includegraphics[width=1\linewidth]{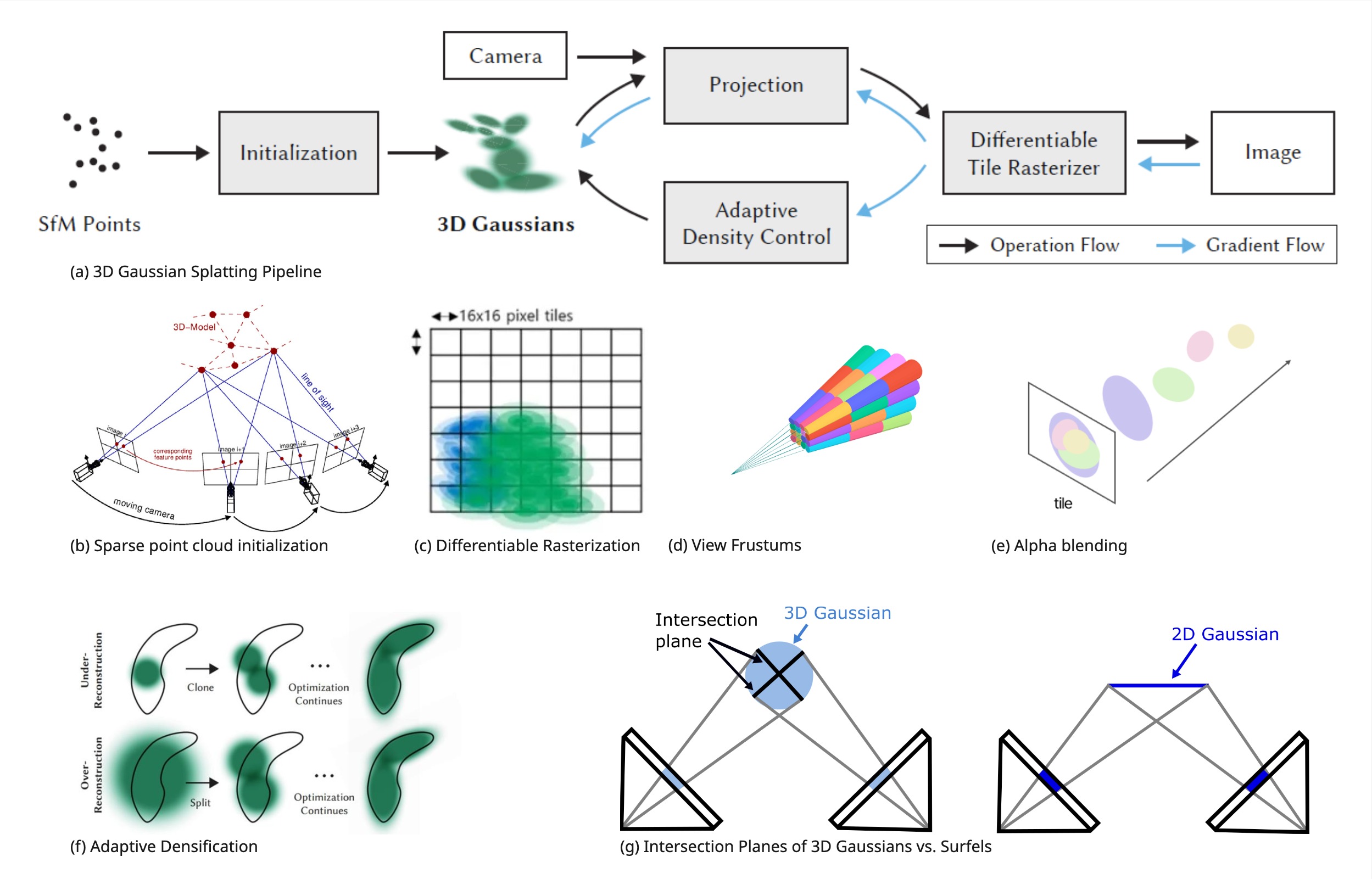}
    \caption{Visualization of key principles in a 3D Gaussian Splatting pipeline (a). Gaussians are initialized from a sparse point cloud (b), for fast rendering the image is split into tiles using differentiable rasterization (c). Projected Gaussians inside a tile's view frustum are sorted by depth (d), this allows $\alpha$-blending to determine the final color of each Gaussian (e). During optimization, adaptive densification controls the number of Gaussians to minimize reconstruction errors (f). View-dependency of color can lead to inconsistency when rendering from different views, flattening the z-scale can improve consistency (g). \\
    Figure compiled from (a)-\cite{Kerbl2023};(b)-\cite{Rahman2023}; (c),(e)-\cite{Han2025};(d)-\cite{Yurkova2023};(f)-\cite{Kerbl2023};(g)-\cite{Huang2024a}}
    \label{fig:3dgs}
\end{figure*}

\subsection{SLAM-based Scene Reconstruction}

\ac{slam} is a fundamental family of techniques in computer vision and robotics that simultaneously estimate a camera’s motion and construct a 3D map of an unknown environment. Early \ac{slam} systems used point clouds,
surfels \cite{Schops2019}, or volumetric representations,
relying on handcrafted feature tracking.
Recent advances introduce differentiable, learning-based representations, most notably 3D Gaussian Splatting \cite{Kerbl2023}. Originally designed for novel view synthesis with known camera parameters, recent works \cite{
Keetha2023, Yan2023} adapt \ac{3dgs} for \ac{slam} by jointly optimizing camera poses and scene geometry via online tracking and incremental mapping. The differentiability of \ac{3dgs} enables gradient-based tracking using geometric (depth accuracy) and photometric (color consistency) loss functions over rendered and observed frames \cite{Yan2023, Keetha2023, 
Sun2024, Zhu2024, Wang2024b, Sun2024a}.
\ac{slam} systems can also benefit from the compactness and continuity of Gaussian primitives improving the memory efficiency. Several works address this topic by compressing Gaussian attributes \cite{Deng2024}, reducing the number of Gaussian primitives by introducing opaque/transparent Gaussians \cite{Peng2024a}, or introducing uncertainty-aware Gaussian fields to balance tracking robustness \cite{Hu2024}. \\
To prevent drift and error accumulation, some systems use keyframe selection with global or hierarchical bundle adjustment framework, ensuring that the estimated camera trajectories remain stable over time \cite{Sun2024a, Ha2024}. Other methods, incorporate \ac{icp} techniques and graph-based optimization, leveraging both front-end and back-end processing for large-scale scene reconstruction \cite{Peng2024a}. While most systems use monocular RGB-D input and adopt optimization techniques, other methods propose the integration of multi-senor data, such as IMU, LiDAR, or stereo data for improved accuracy and camera pose propagation \cite{Lang2024, Sun2024a}.\\
Despite these commonalities, differences emerge in the mapping strategies. They vary from coarse-to-fine refinement using sparse pixel initialization and depth-based updates, to silhouette-guided rendering that incrementally refines poses, to dividing the scene into sub-maps for localized optimization \cite{Yan2023, Keetha2023}. \\
Going beyond static mapping, newer semantic \ac{slam} systems incorporate semantic features directly into Gaussian parameters. This enables joint optimization of semantics, geometry, and appearance in a unified framework \cite{Ji2024, Li2024f, Zhu2024}. In parallel, dynamic \ac{slam} systems adapt \ac{3dgs} to model non-rigid environment such as deformable tissues in endoscopy. These systems integrate temporal deformation fields, occlusion-aware loss function, and surface-aligned regularization to capture motion while maintaining consistent tracking \cite{Wang2024b, Zhu2024a, Huang2024c}. \\
\ac{slam} in unconstrained environments often deals with sparse observations, occlusion, or ambiguous input. Recent work proposes rendering-guided densification to refine geometry by injecting new Gaussians in underrepresented areas \cite{Sun2024}. Other methods reduce the number of splats by modeling scenes with hybrid Gaussians that selectively capture solid vs. transparent structures \cite{Peng2024a}.
}

\subsection{Sparse-View and Single-Image Scene Reconstruction}

\renewcommand{\arraystretch}{1.2}
\begin{table*}[h]
\caption{Summary of object-level single-view 3D scene reconstruction and modeling methods}
\centering
{
\begin{tabular}{@{} |l|c|c|c|c| @{}}
\hline
\textbf{Method} & \textbf{Framework} & \textbf{Model architecture} & \textbf{3D representation} & \textbf{Learning strategy}  \\
\hline
\multicolumn{5}{|l|}{\textbf{3D scene reconstruction}} \\
\hline
Total3D~\cite{nie2020total3dunderstanding}  &end-to-end &2D CNNs & template mesh  &supervised  \\
IM3D~\cite{zhang2021holistic}  &end-to-end & 2D CNNs, GCN &SDF &supervised   \\
InstPIFu~\cite{liu2022towards}  &end-to-end &2D CNNs & implicit function &supervised  \\
SSR~\cite{chen2024single}  &end-to-end &2D CNNs &SDF &supervised  \\
Gen3DSR~\cite{dogaru2024generalizable}  &modular &FMs &neural field &zero-shot  \\
MIDI~\cite{huang2024midi}  &end-to-end &TFs &latent feat. &diffusion   \\
DPA~\cite{zhou2024zero}  &modular &FMs &mesh &zero-shot  \\
CAST~\cite{yao2025cast}  &modular &FMs &latent feat. &diffusion  \\
\hline
\multicolumn{5}{|l|}{\textbf{3D scene modeling}} \\
\hline
IM2CAD~\cite{izadinia2017im2cad}  & modular &2D CNNs &- &supervised \\
Mask2CAD~\cite{kuo2020mask2cad}  &end-to-end &2D CNNs &- &supervised  \\
Patch2CAD~\cite{kuo2021patch2cad}  &end-to-end &2D CNNs & - &supervised \\
ROCA~\cite{gumeli2022roca}  &end-to-end & 2D\&3D CNNs & NOCs &supervised \\
PSDR-Room~\cite{yan2023psdr}  &modular &FMs &PC &zero-shot  \\
DiffCAD~\cite{gao2023diffcad}  &modular &FMs &PC \& NOCs &diffusion  \\
Digital Cousin~\cite{dai2024automated}  &modular &FMs &PC &zero-shot  \\
Diorama~\cite{wu2024diorama}  &modular &FMs &PC &zero-shot \\
\hline
\end{tabular}
}
\label{tab:scene_reconstruction}
\end{table*}

Reconstructing 3D scenes or objects from only one or a few images has become an active research frontier in digital twin generation~\cite{fan2024instantsplatsparseviewsfmfreegaussian, dai2024automated}. 
Unlike traditional multi-view pipelines that rely on dense, overlapping captures and precise camera poses~\cite{schonberger2016structure}, sparse-view and single-image reconstruction must infer missing geometry, texture, and lighting from extremely limited visual evidence. 
This underconstrained problem demands strong structural priors, semantic reasoning, or generative diffusion guidance to produce plausible, view-consistent 3D scenes.

\subsubsection{Optimization-based Sparse Reconstruction} 
Several recent frameworks extend 3D Gaussian Splatting (3DGS)~\cite{Kerbl2023} to sparse-input settings. 
Optimization-based methods~\cite{Paliwal2024, Yang2024} iteratively reconstruct 3D Gaussians guided by additional priors such as monocular depth, multi-view flow, or visual hulls. These regularization terms enforce view-consistent geometry and can be optimized through differentiable rendering. 
Next-generation frameworks, such as InstantSpat~\cite{fan2024instantsplatsparseviewsfmfreegaussian} and Dust to Tower~\cite{Cai2024a}, reconstruct scenes from sparse-view images without explicit \ac{sfm}. 
They leverage dense stereo correspondence models such as DUSt3R~\cite{wang2024dust3r} and MASt3R~\cite{leroy2024grounding} to predict pixel-aligned depth and initialize camera parameters. 
By constructing co-visibility graphs and optimizing 3D Gaussian parameters jointly with camera alignment via photometric error minimization, they achieve consistent reconstructions even with minimal input. 

\subsubsection{Amortized Feed-Forward Inference}
Complementary to optimization, amortized models~\cite{Xu2024c, Xu2024, Szymanowicz2023}
learn to directly infer 3D Gaussian parameters from one or more RGB images using convolutional or transformer-based architectures. 
They typically condition on camera poses or rays and employ scan-ordered sequential or tokenized network designs to efficiently scale to thousands of Gaussians while preserving detail. 
Fang et al.~\cite{fang2025single} propose SVG3D, which integrates RGB and monocular depth in a U-Net to predict parameters of 3D Gaussian ellipsoids. 
These feed-forward methods provide real-time reconstruction capabilities and generalize well to unseen configurations, making them promising for online digital twin updates.

\subsubsection{Generative Priors and Diffusion-Based Guidance} 
Generative priors can compensate for missing views by synthesizing plausible geometry and texture. 
Hybrid frameworks combine explicit Gaussian reconstruction with 2D diffusion-based refinement for occlusion completion, appearance refinement, and even editing~\cite{Wu2024a, Chen2023}. 
These methods start from a coarse 3D reconstruction, render multi-view images, refine them through diffusion-based image enhancement, and re-optimize the 3D scene to improve fidelity, even from a single image~\cite{Zhang2024, Chen2023, Wu2024a}. 
Diffusion-based 3D generation models further generalize this paradigm. For instance, GSD~\cite{mu2024gsd} denoises random Gaussian clouds while enforcing consistency between rendered and observed views. Zero-1-to-3~\cite{liu2023zero} learns a viewpoint-conditioned image diffusion model for novel-view synthesis and 3D shape estimation. Basak et al.~\cite{basak2024enhancing} introduce a hybrid 2D-3D difusion framework, in which a 3D diffusion prior captures global geometry, while a 2D diffusion model refines texture and appearance. Similarly, other hybrid diffusion pipelines such as PC$^2$~\cite{melas2023pc2} and CDM~\cite{jiang2025consistency} couple 3D structural priors with 2D image features for spatially coherent denoising. Domain-specific extensions include PSHead~\cite{yang2025pshead} for single-image human head reconstruction and SINGAPO~\cite{liu2024singapo}, which models articulated household objects (e.g., refrigerators, cabinets) through part-wise diffusion.

\subsubsection{Asset Retrieval and Digital Cousins}
A complementary strategy retrieves or assembles existing 3D assets instead of generating them from scratch. 
Digital Cousins~\cite{dai2024automated} are synthetic 3D scenes generated from a single RGB image that do not exactly replicate real-world assets but retain key geometric and semantic information.
Objects are detected using open-vocabulary models~\cite{melnik2023uniteam, yenamandra2024towards} and segmented with GroundedSAM-v2. Depth information is inferred with mono-depth estimators such as Depth Anything V2~\cite{yang2025depth}. Segmentation results are matched to 3D assets (e.g., BEHAVIOR-1K~\cite{li2022behavior}) using CLIP scores and DINOv2~\cite{oquab2023dinov2} embeddings. 
Assets are placed into physically consistent layouts through geometry refinement. 
CAST~\cite{yao2025cast} enhances Digital Cousins by generating high-quality meshes with physics-aware correction, rather than matching segmentations to an object library. Diorama~\cite{wu2024diorama} uses hierarchical retrieval combining visual and textual similarity via DuoDuoCLIP~\cite{lee2024duoduo}. 
Earlier approaches such as Im2CAD~\cite{izadinia2017im2cad}, Mask2CAD~\cite{kuo2020mask2cad}, Patch2CAD~\cite{kuo2021patch2cad}, and ROCA~\cite{gumeli2022roca} and DiffCAD \cite{gao2024diffcad} align detected objects to CAD models but require extensive image–CAD annotations. 
Recent zero-shot methods~\cite{dai2024automated, wu2024diorama} overcome this limitation through visual foundation models and \ac{llm} reasoning.
AdaptiveCLIP~\cite{song2025adaptive} embeds images and 3D models into a shared feature space using CLIP for open-domain shape retrieval.

\subsubsection{Trends and Implications}
Overall, sparse-view reconstruction in digital twins is evolving toward hybrid frameworks that integrate explicit geometry, neural inference, and generative priors. 
Optimization-based 3DGS systems ensure geometric consistency, while diffusion or retrieval models inject realistic appearance and semantic plausibility. 
These approaches collectively reduce the requirements on sensors and capture setups while maintaining functional fidelity, enabling scalable and automated digital twin generation from minimal visual input. They feed directly into generative 3D synthesis approaches~\cite{poole2023dreamfusion} and demonstrate that powerful 2D priors can supervise 3D generation without explicit 3D data, opening the path toward text- or image-conditioned 3D creation.

\section{Input Modalities and Acquisition Challenges}

\subsection{2D Monocular Video}
\sh{
Reconstructing 3D scenes from 2D videos generally requires multi-view captures, but even single-view sequences can contain multi-view-like cues depending on camera motion and scene structure \cite{Gao2022}.

\subsubsection{Motion Blur, Defocus, Downsampling Blur}
Blur and other degradations reduce \ac{sfm} accuracy and reconstruction quality, especially with unreliable pose estimates. Recent methods explicitly model blur for better alignment between rendered and observed frames: simulating blur via sub-frame rendering and averaging \cite{Oh2024}, synthesizing blurred frames from interpolated sharp renderings along camera trajectories \cite{Zhao2024b}, or predicting spatially varying per-pixel blur kernels through a Blur Proposal Network refined in a coarse-to-fine manner \cite{Peng2024}.
}

\subsubsection{Wide-Angle Lenses}
\am{
Wide-angle lenses introduce strong distortion, producing noisy 3DGS reconstructions. A self-calibrating resampling framework \cite{deng2025self} addresses this by combining invertible residual networks \cite{behrmann2019invertible} with explicit grids.
}

\subsubsection{Rolling Shutter Cameras}
\sh{
Rolling shutter sensors capture images line by line, introducing motion-induced distortions that violate global-pose assumptions. To mitigate this, \cite{Seiskari2024} models time-varying camera poses during exposure, accurately simulating rolling-shutter effects.
}

\subsubsection{Appearance-Conditioned Gaussians}
\sh{
To achieve realistic reconstructions under varying illumination or transient occlusions, \cite{Dahmani2024} learns image-dependent appearance codes that adapt color per view and applies image-specific opacity to suppress inconsistent elements like moving objects.
}

\subsubsection{Non-Calibrated Cameras}
\sh{
Non-calibrated cameras lack known intrinsic or extrinsic parameters, making 2D–3D mapping and \ac{sfm} unreliable, especially with sparse or textureless data \cite{Fu2023}. Two main strategies address this: (1) \textit{pose refinement}, which optimizes initial pose estimates from COLMAP or similar sources via render-and-compare or 2D–3D feature alignment \cite{Liu2024e, Sun2023}, and (2) \textit{pose-free reconstruction}, jointly estimating geometry and poses from unposed images using pixel-aligned point maps, depth refinement, or pose networks \cite{Fu2023, Kang2024, Fan2024b, Cai2024a}. Vision transformer-based initializations further scale training across unordered datasets \cite{Chen2024d, Shi2025}, enabling differentiable pipelines that recover geometry and poses jointly, achieving high rendering and pose accuracy without precomputed camera input \cite{Li2024i}.
}

\subsection{LiDAR and RGB-D Inputs}
\sh{
In many systems depth data is utilized through depth-aware loss terms or direct \ac{icp} alignment \cite{Ha2024, Sun2024a, Yan2023}. Additionally it can be utilized for more precise surface modeling and 3D point placement or constrain 3D structures across different views \cite{Peng2024a, Sun2024, Li2024f, Zhu2024}.
A major drawback, however, is the short effective range of depth sensors. LiDAR sensors on the other hand use one or more laser sensors to measure a scene by bouncing laser beam off of objects. 
LiDAR points together with RGB data for color can directly be used to initialize 3D point-based representations for environments where visual features are sparse or unreliable \cite{Lang2024}. 
}

\section{Light and reflections}
\label{light}

\subsection{Relighting}

\begin{figure}
\centering
\includegraphics[width=\linewidth]{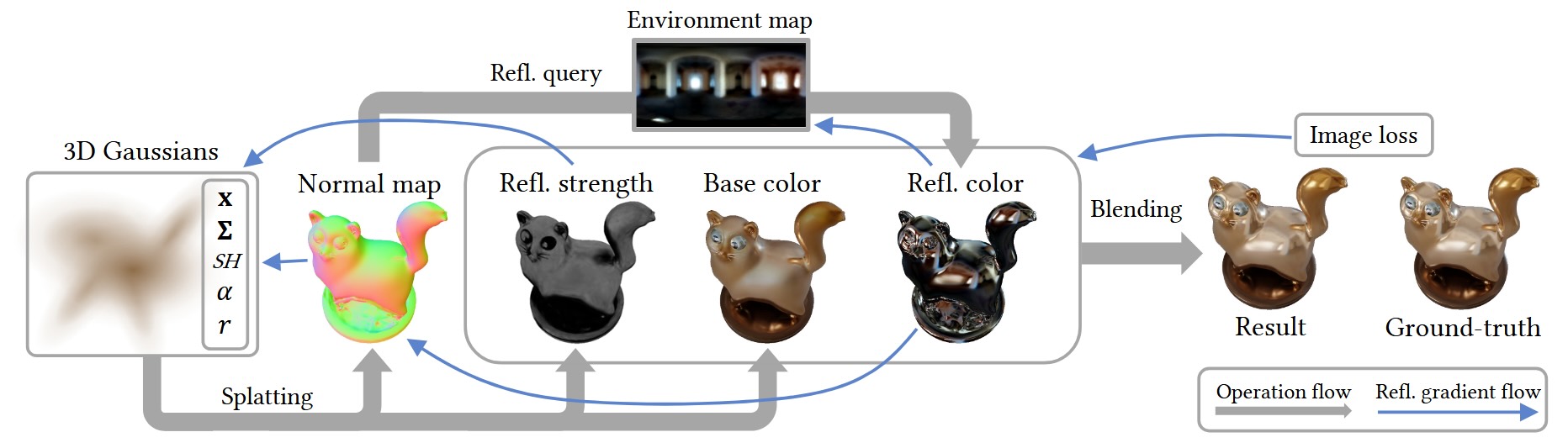}
\caption{Example rendering pipeline for Gaussian splatting with reflections~\cite{ye20243DGaussianSplatting}. The pipeline is similar in most of the approaches.}
\label{fig:relight}
\end{figure}

\ms{
Standard \ac{3dgs} uses spherical harmonics for view-dependent color modeling, but the baked-in lighting limits relighting capabilities. To overcome this, recent works extend the rendering pipeline with a Bidirectional Scattering Distribution Function (BSDF) model \cite{R3DG2023,liang2023gs,jiang2023gaussianshader}. This adds learnable parameters such as albedo, roughness, and normals, while using a global texture for environment lighting that can be modified at render time (see Fig.~\ref{fig:relight}).

Further improvements include indirect lighting simulation using single-bounce ray tracing with global textures or per-Gaussian spherical harmonics \cite{shi2023gir}, and deferred rendering pipelines \cite{wu2024deferred,ye20243DGaussianSplatting,chen2024gigs}, which enable accurate per-pixel shading, reflections, and explicit light sources \cite{bi2024rgs}. Similar BSDF-based methods are also applied in 2DGS \cite{Huang2024a}, with added surface regularization \cite{du2024gsid} and local mesh reconstruction for improved surface quality \cite{yao2025reflective}. Multi-light training enhances lighting generalization, from per-image point lights \cite{bi2024rgs} to multi-environment setups \cite{li2024recap}.

Besides explicit BSDF approaches, implicit relighting via latent per-Gaussian features has been proposed \cite{fan2024rng}, where light and view directions condition the decoding process. Table~\ref{tab:relight} summarizes the main characteristics of these methods.
}

\begin{table*}[ht!]
    \caption{Comparison of main features of different relighting approaches (values for shadows column: RT - ray-traced, AO - baked ambient occlusion) }
    \centering
    \begin{tabular}{|l|c|c|c|c|}
        \hline
Method                                        & Dynamic lights & Shadows & Rendering & Representation\\
\hline
Relightable \ac{3dgs} \cite{R3DG2023}              & no             & RT      & forward   & explicit\\
GS-IR \cite{liang2023gs}                      & yes            & RT+AO   & forward   & explicit\\
GaussianShader \cite{jiang2023gaussianshader} & no             & no      & forward   & explicit\\
GIR \cite{shi2023gir}                         & no             & RT      & forward   & explicit\\
DeferredGS \cite{wu2024deferred}              & no             & no      & deferred  & explicit\\
3DGS-DR \cite{ye20243DGaussianSplatting}      & no             & no      & deferred  & explicit\\
GI-GS \cite{chen2024gigs}                     & no             & RT+AO   & deferred  & explicit\\
GS$^3$ \cite{bi2024rgs}                       & yes            & RT      & deferred  & explicit\\
GS-ID \cite{du2024gsid}                       & yes            & AO      & deferred  & explicit\\
Reflective GS \cite{yao2025reflective}        & no             & RT      & deferred  & explicit\\
ReCap \cite{li2024recap}                      & no             & no      & forward   & explicit\\
RNG \cite{fan2024rng}                         & yes            & RT      & deferred  & implicit\\
\hline
    \end{tabular}
    \label{tab:relight}
\end{table*}

\subsection{Mirrors}

\begin{figure*}[h!]
\centering
a) \includegraphics[width=0.3\linewidth]{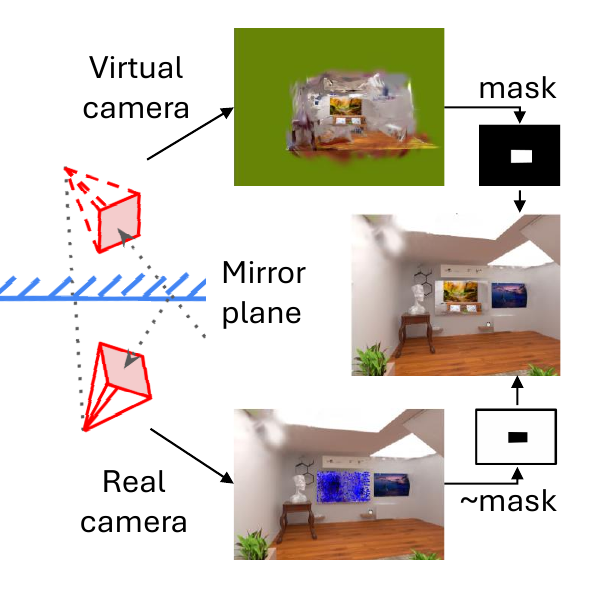}
b) \includegraphics[width=0.25\linewidth]{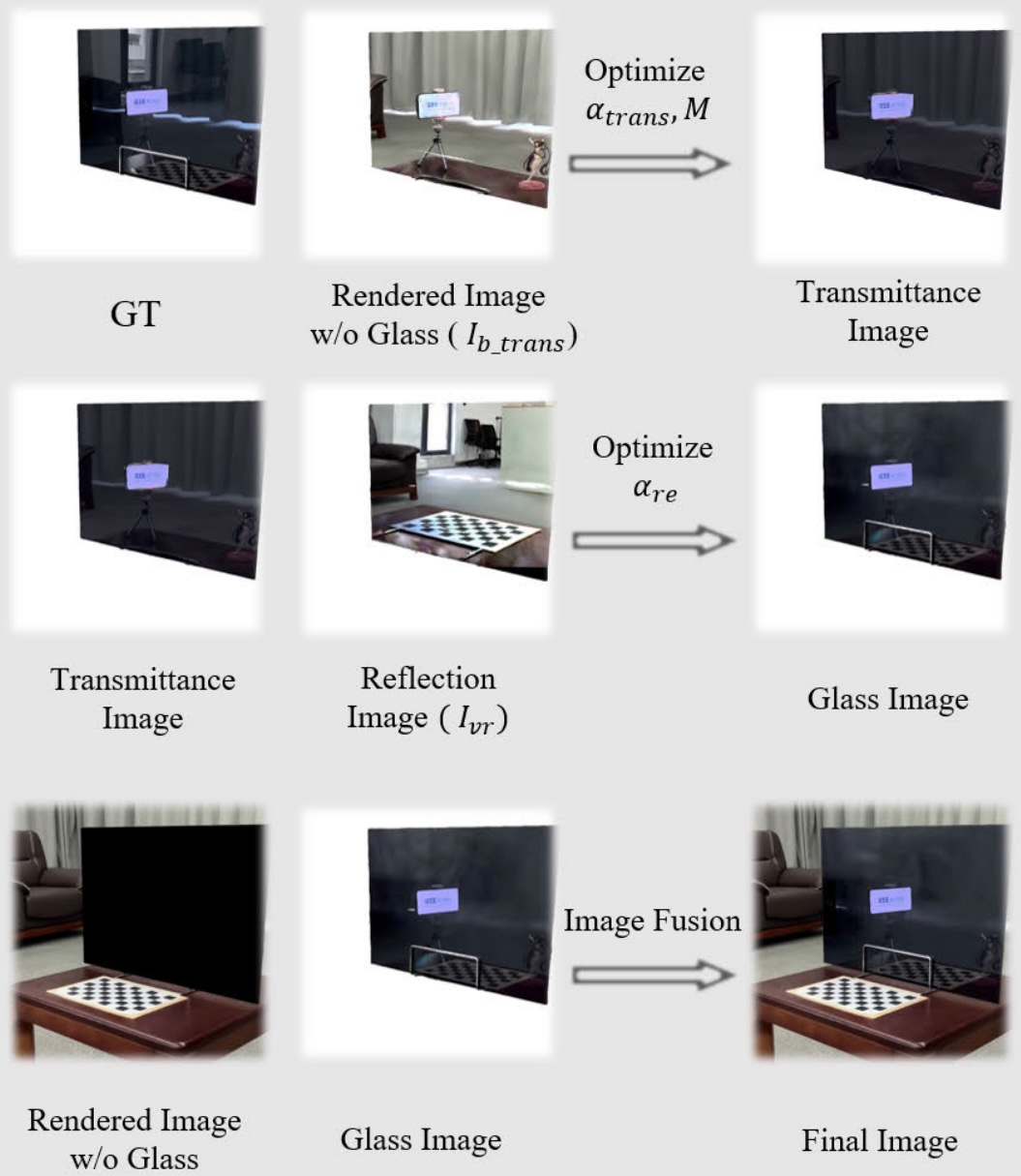}
c) \includegraphics[width=0.3\linewidth]{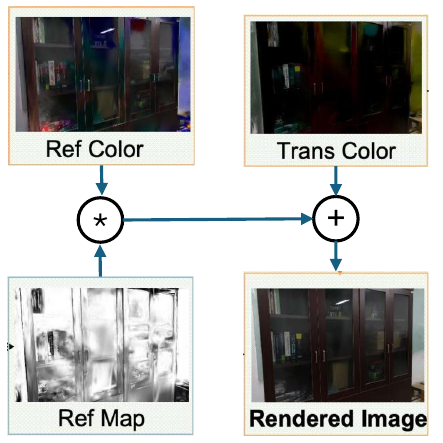}
\caption{Three levels of mirror rendering: a) the perfect mirror \cite{wang2024gaussianmirrors}, b) semi-transparent glass \cite{cao_glassgaussian_2025}, c) non-uniform reflection strength \cite{zhang2024refgaussian}}
\label{fig:mirror}
\end{figure*}

\ms{
While existing methods model glossy surfaces well, mirrors remain challenging.
In standard \ac{3dgs}, mirrors are often treated as separate rooms or result in blurred reflections encoded in color harmonics.

Recent works address this by detecting mirror regions and rendering reflections from virtual cameras positioned behind the mirror plane \cite{meng2024mirror3dgs,wang2024gaussianmirrors,liu2024mirrorgaussian}, as shown in Fig.~\ref{fig:mirror}. Mirror masks are either manually provided or automatically detected \cite{kirillov2023segment}. The mirror plane is derived from \ac{sfm} point clouds \cite{liu2024mirrorgaussian} or from masked Gaussian initialization \cite{meng2024mirror3dgs}, followed by reflection refinement through plane or virtual camera adjustments \cite{meng2024mirror3dgs,wang2024gaussianmirrors}.

GlassGaussian \cite{cao_glassgaussian_2025} extends these methods to semi-transparent surfaces by modeling angle-dependent reflection via spherical harmonics and rendering transparency and reflection in separate passes. These approaches support dynamic scenes since reflections are rendered explicitly, unlike methods relying on static color encoding \cite{zhang2024refgaussian,wang2024space}.
}

\am{

\subsection{Open Challenges}

Photorealistic digital twins depend on accurate light transport simulation, yet current methods treat diffuse, specular, and transparent effects separately. While BSDF-based relighting and mirror reconstruction have advanced, a unified, physically consistent framework remains missing. \ac{3dgs} offers an efficient middle ground between NeRF realism and practical performance, but a major gap persists: translating \ac{3dgs} into industry-standard formats (USD, glTF, meshes) without losing fidelity. This conversion challenge limits real-world deployment in game engines, simulations, and manufacturing pipelines.

Key research priorities include:
\begin{enumerate}
    \item \textbf{Robust conversion tools} preserving appearance and lighting across standard formats.
    \item \textbf{Hybrid frameworks} that jointly optimize \ac{3dgs} and mesh/CAD proxies.
    \item \textbf{Physically grounded optimization} enforcing material and energy consistency.
    \item \textbf{Benchmark datasets} with ground-truth materials and lighting for quantitative evaluation.
\end{enumerate}

}

\section{Temporal Dynamics}
\label{dynamics}

\subsection{Scene-Level Temporal Dynamics}

\begin{figure*}[t]
\centering
\includegraphics[width=1.0\linewidth]{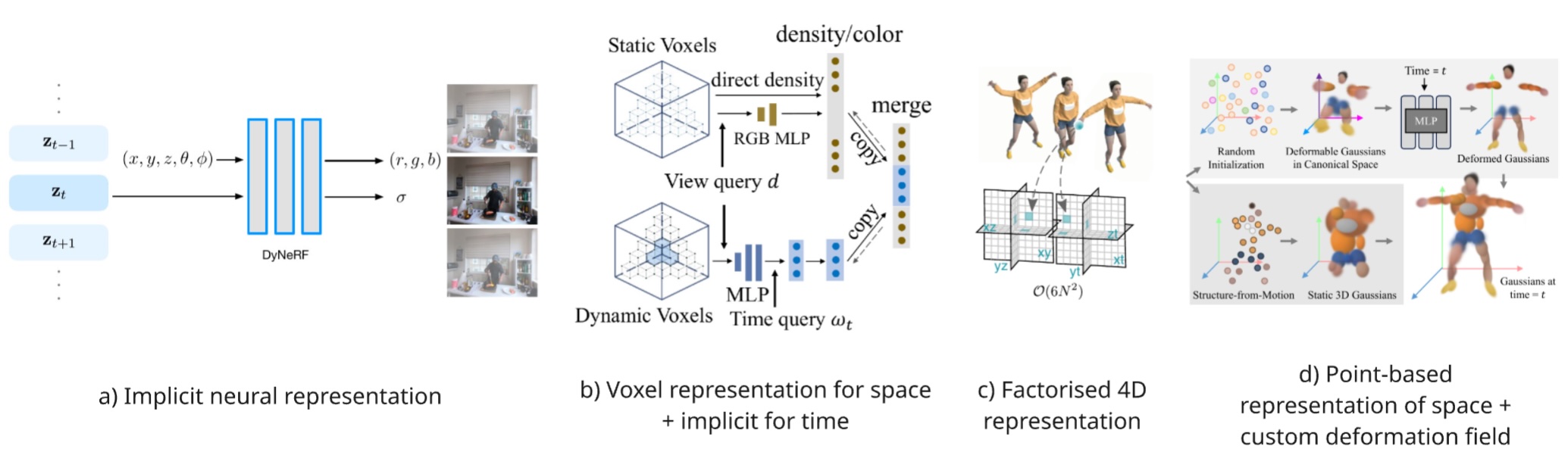}
\caption{Evolution of representation methods for dynamic content utilizing differentiable volumetric rendering. The concepts involve fully-implicit representation by neural networks (a), voxel-based 3D representation supported by implicit dynamics (b), fully 4D voxel-based representation using factorization for tractability (c), point-based representation of space (typically by Gaussians) with different methods for representation point dynamics (d). Figure compiled from \cite{li22neural3d,wang23mixedneural,fridovich23kplanes}}
\label{fig:temporal_representation}
\end{figure*}

\begin{aw}
In simple cases like rigid object manipulation, scene dynamics can be modeled as rotation and translation \cite{wen2023bundlesdf}
For free motion, more complex representations are needed, as summarized in Fig.~\ref{fig:temporal_representation}.

\subsubsection{Implicit Neural Representations}
Implicit methods use neural networks to encode both spatial and temporal structures. In \cite{li22neural3d}, a single MLP encodes spatial and temporal data with learnable time embeddings, while \cite{pumarola21dnerf} introduces a separate deformation network to map temporal changes to a static canonical space.

\subsubsection{Voxel Grid Representations}
Explicit voxel-based methods reduce memory costs by combining voxel grids with deformation or radiance networks. \cite{fang22tineuvox} encodes motion information in sereral ways: small neural network, voxel grid and a small radiance network processing grid features and time component. \cite{gan24v4d} skips explicit deformation, embedding time into features from density and texture volumes before radiance computation.

\subsubsection{Planar Factorizations of Voxel Grids}
Spatio-temporal grids can be efficiently factorized into 2D planes (e.g., XY, XT, YZ) to reduce memory usage \cite{fridovich23kplanes,cao23hexplane}. Each plane stores either feature vectors \cite{fridovich23kplanes} or basis vector coefficients \cite{cao23hexplane}. These decompositions can support rendering methods like Depth Peeling \cite{xu24.4k4d} or splatting \cite{Wu2023}. K-plane models can be used to represent dynamic Gaussian deformation via learnable features processed through MLPs \cite{Wu2023}.

\subsubsection{Point-based Representations}
Point-based methods represent the scene as unstructured sets of Gaussians or surfels and model temporal dynamics through deformation or trajectory-based motion fields \cite{Yang2023a, Li2023, Luiten2023, Yang2023, xu24.4k4d,kratimenos24dynmf,huang24scgs, Wu2024, Wang2024d,  zhang2024egogaussian, Bae2024, Lu2024, Lei2024a, Sun2024b, Shaw2023}. 
Deformation can be coordinate-based using MLPs \cite{Yang2023a}, voxel-encoded \cite{Sun2024b}, or Gaussian-based with per-particle embeddings \cite{Bae2024} or learner via geometry-aware networks \cite{Lu2024}. Trajectories can be explicit \cite{Luiten2023,Lei2024a} or learned \cite{kratimenos24dynmf}, often defined through control points \cite{huang24scgs,Wu2024} or bones \cite{Wang2024d}. Some methods extend this to full 4D Gaussians representing space-time volumes \cite{Yang2023}. 

\subsubsection{Decoupling of Static and Temporal Information}
Many models separate static canonical structures from dynamic motion using deformation fields, while others unify them into a single model \cite{li22neural3d} or optimize motion per frame \cite{Luiten2023}. Fully 4D Gaussians \cite{Yang2023} directly encode time, while multi-sampling of canonical configurations in time improves flexibility \cite{Shaw2023}.

\subsection{Regularization of Spatio-Temporal Patterns}

\subsubsection{Regularization by Representation}
Structured representations, such as K-Planes \cite{fridovich23kplanes,Wu2023} or Hex-Planes \cite{cao23hexplane,Yin2023,Ren2023}, inherently enforce correlations between spatial and temporal voxels. Similarly, 4D Gaussian models \cite{Yang2023} promote coherent motion through coarse-grained spatio-temporal units.

Point-based models also impose regularization through shared motion bases or template trajectories \cite{kratimenos24dynmf}
, often organized via graphs \cite{Lei2024a} or control points for articulated motion \cite{huang24scgs,Wu2024}.

\subsubsection{Regularization Loss Functions}
Smoothness is often enforced by tailored loss functions. In K-Planes, spatial total variation and temporal smoothness losses prevent abrupt changes \cite{fridovich23kplanes}. Gaussian-based methods add constraints for local rigidity, rotation similarity, and maintenance of long-term isometry \cite{Luiten2023,Bae2024}.

Motion representations using limited trajectory bases or control points \cite{kratimenos24dynmf,huang24scgs,Wu2024} are further regularized with sparsity and rigidity losses, often based on ARAP (As Rigid As Possible) principles \cite{Lei2024a,huang24scgs}. Optical flow–based losses also align 3D and 2D motion cues 
\cite{Gao2024}.

\subsubsection{Regularization by Foundation Models}
Foundation models such as diffusion models support 4D scene reconstruction by providing priors for motion and appearance \cite{Yin2023,Ren2023,Pan2024,Gao2024,Lei2024a,Wu2024}. Techniques like Score Distillation Sampling (SDS) use pretrained models to supervise spatial consistency via image diffusion \cite{Yin2023,Gao2024, Wu2024} or video diffusion for spatio-temporal coherence \cite{Ren2023,Pan2024}.

\subsection{Efficiency Considerations of Temporal Methods}
NeRF-based temporal models are computationally expensive, requiring days for training and seconds per-frame inference \cite{pumarola21dnerf,li22neural3d}. Voxelized or hybrid representations \cite{fang22tineuvox,gan24v4d,wang23mixedneural} significantly reduce training times to under an hour and enable near-real-time inference. Factorized voxel systems like K-Planes and Hex-Planes further improve efficiency \cite{fridovich23kplanes,cao23hexplane}, even achieving real-time performance with depth-peeling renderers \cite{xu24.4k4d}.

Gaussian Splatting methods achieve sub-hour training and real-time rendering, often surpassing 100 fps \cite{Yang2023a,Wu2023,Li2023,kratimenos24dynmf}.

\subsection{Applicability of Temporal Methods in Robotics}

Gaussian-based methods \cite{Yang2023a,Wu2023,Li2023,kratimenos24dynmf} achieve high-speed training and inference, suitable for real-time mapping. Monocular datasets like DNeRF \cite{pumarola21dnerf} are commonly used to test such methods \cite{Yang2023a,huang24scgs,kratimenos24dynmf,Yang2023,Wu2023}.

Multi-camera methods include \cite{Luiten2023,xu24.4k4d,Li2023}, while monocular approaches rely on motion regularization via control points \cite{huang24scgs}, shared trajectories \cite{kratimenos24dynmf,Lei2024a}, 4D Gaussians \cite{Yang2023}, deformation factorization \cite{Wu2023}, optical flow \cite{Gao2024}, or diffusion priors \cite{Pan2024}.

Point tracking is naturally supported by Gaussian-based structures, with additional tracking regularization in \cite{Luiten2023,kratimenos24dynmf,huang24scgs}. Some methods also model opacity changes for dynamic appearances \cite{Li2023}.

Task-specific methods such as 
\cite{zhang2024egogaussian} perform joint 6DoF tracking and reconstruction, distinguishing rigid object motion from static backgrounds.
\end{aw}

\am{
\subsection{Open Challenges}

Extending \ac{3dgs} to dynamic 4D scenes enables real-time, temporally coherent reconstructions, but introduces key challenges: maintaining long-sequence temporal consistency, disentangling geometry from motion in monocular settings, learning motion patterns and efficiently representing articulated dynamics. While factorized spatiotemporal representations and Gaussian-based deformation fields improve efficiency, scaling to extended sequences and streaming reconstruction remains an open problem.  

Research and development targets include:
\begin{enumerate}
    \item Standardized temporal encoding for USD/glTF and streaming-friendly compression.
    \item Conversion pipelines translating 4D Gaussians to skeletal animation or morph targets.
    \item Physics-aware temporal priors and joint optimization of geometry, appearance, and dynamics.
    \item Learned priors and differentiable constraints for articulated motion and semantics.
    \item Incremental, memory-efficient online reconstruction for embodied AI and real-time robotics.
\end{enumerate}

Addressing these challenges will bridge the 4D representation gap, enabling \ac{3dgs}-based digital twins to transition to practical systems for VFX, gaming, robotics, and immersive applications.
}

\section{Physical Properties}
\label{physical}

\gn{
Extracting physical properties from videos is essential for accurate physics-based simulations in digital twins \cite{melnik2023benchmarks}. This section reviews methods that estimate physical parameters—such as mass, inertia, friction, and viscosity—from video data for integration into physics-aware 3D or Gaussian-based scene representations.
}

\subsection{Physics Simulations for Digital Twins}

\gn{
Modern physics simulators have become faster, more reliable, and diverse, supporting robotics, industrial, and medical applications. This section outlines leading simulation platforms and scene description formats relevant to digital twin creation.
}

\subsubsection{State-of-the-art Simulation Technologies}

\gn{
According to \cite{kaup2024review}, top simulation engines for robotics include Isaac Sim, MuJoCo, Gazebo, PyBullet and Webots, which primarily simulate rigid and multi-joint systems in URDF and MJCF formats. These engines are optimized for reinforcement learning and continuous control.

For soft-body and medical simulations, SOFA \cite{Faure2011SOFAAM} uses the Finite Element Method (FEM) for modeling deformable materials, enabling tasks such as cutting and robotic manipulation \cite{haiderbhai2024simulating}. It uses scene graphs but does not support URDF or MJCF.

ProjectChrono \cite{pr9101813} focuses on particle and granular dynamics, employing the Discrete Element Method (DEM) for GPU-accelerated vehicle and terrain simulations, using JSON scenes with URDF parsing.

Photorealistic simulation engines like Isaac Sim, Unreal Engine, Unity, Godot, and Blender provide high-quality rendering for robotics and visualization. Isaac Sim integrates PhysX for robotics simulation, while Unreal Engine and Unity support plugins such as CARLA \cite{Dosovitskiy17} and URoboSim \cite{urobosim21cvpr}. Unity also supports MuJoCo integration and USD scene imports.

Genesis \cite{authors2024genesis} combines multiple solvers into a unified framework with modular generative components. The choice of simulator depends on the application—photo-realistic rendering benefits perception studies, while dynamic control tasks require engines supporting direct physics manipulation. Complex operations like cutting or pouring depend on FEM or DEM modeling.
}

\subsubsection{Scene Description Formats}

\gn{
Simulation environments rely on scene description formats to define objects and dynamics. Common formats include MJCF (MuJoCo), SDF (Gazebo), and PROTO (Webots) (see Fig.~\ref{fig:SceneDescription}). However, incompatibility between formats can hinder interoperability. URDF is widely used for rigid and multi-joint systems \cite{tola2024understanding}, though it lacks features for soft-body, fluid, or photorealistic simulations.

\begin{figure}[h]
\centering
\includegraphics[width=\linewidth]{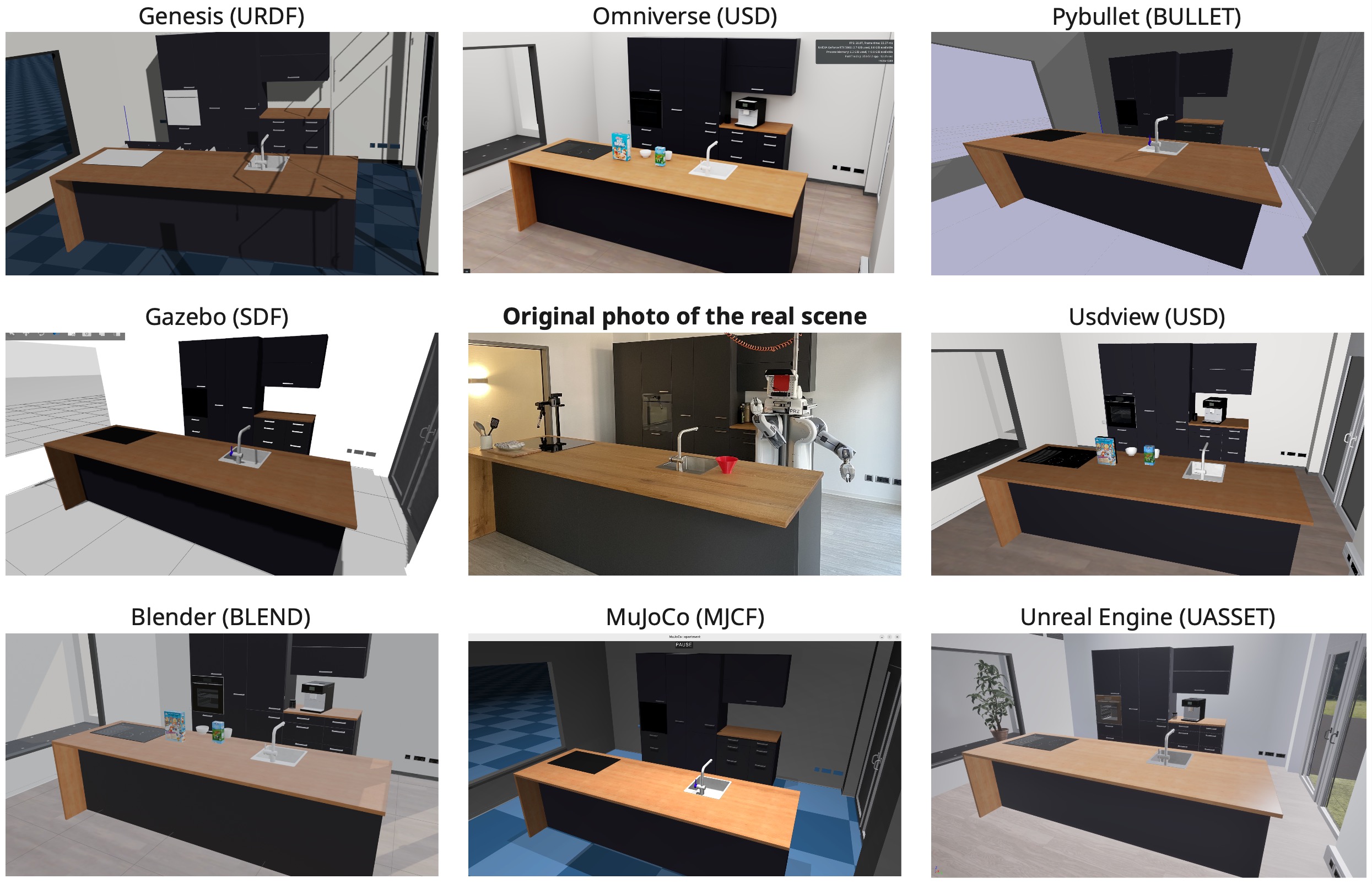}
\caption{Different simulators and their scene descriptions, generated from a real environment.}
\label{fig:SceneDescription}
\end{figure}

To address these limitations, the Universal Scene Description (USD) \cite{elkoura2019deep} developed by Pixar enables flexible, extensible scene encoding. Supported by NVIDIA Omniverse and Isaac Sim, USD can represent rigid, soft, and fluid bodies while incorporating rendering and physical properties. It serves as a unified format with numerous converters, fostering research in knowledge graphs \cite{nguyen2024translating}, annotated scene datasets, and neuro-symbolic 3D scene representations \cite{hossain2023neuro}.
}

\subsection{Learning Physics-Based Attributes from Videos}

\gn{
Different simulators support distinct physics properties and scene formats (see Table~\ref{tab:simulators-comparison}). Once physical parameters are estimated from video data, they can be encoded into scene descriptions for direct simulation.
}

\begin{table*}[h]
\caption{Comparison of Simulators for Digital Twins}
\centering
\begin{tabular}{|p{1.7cm}|p{2cm}p{1.4cm}|p{12cm}|}
\cline{1-4}
\multirow{2}{*}{} & \multicolumn{2}{c|}{\begin{tabular}[c]{@{}c@{}}Scene Description \\ Formats\end{tabular}} & \multicolumn{1}{c|}{\multirow{4}{*}{Outstanding features}} \\ \cline{2-3}
                  & \multicolumn{1}{l|}{Main} & \begin{tabular}[c]{@{}l@{}}Compatible\end{tabular} &                                 \\ \cline{1-4}
MuJoCo          & \multicolumn{1}{l|}{MJCF}    & URDF                                                 & Fast and accurate physics for continuous control with smooth contact dynamics; great for reinforcement learning; used in research; Url: \url{https://mujoco.org} \\ \cline{1-4}
Gazebo          & \multicolumn{1}{l|}{SDF}     & URDF                                                 & Modular and ROS-integrated; supports realistic sensor simulation; suitable for real-world robot prototyping; large plugin support; Url: \url{https://gazebosim.org} \\ \cline{1-4}
PyBullet    & \multicolumn{1}{l|}{BULLET}  & SDF URDF MJCF & Easy scene prototyping with ML support; integrates OpenGL rendering for lightweight digital twin visualization; Url: \url{https://pybullet.org}\\ \cline{1-4}
Webots         & \multicolumn{1}{l|}{PROTO}   & URDF                                                 & Fast setup of indoor environments with GUI and rich built-in robot/sensor models; User-friendly GUI; good for education; Url: \url{https://cyberbotics.com} \\ \cline{1-4}
Drake             & \multicolumn{1}{l|}{MJCF}    & URDF                                                 & Analytical control modeling with structured dynamics; great for optimization in constrained spaces; Url: \url{https://drake.mit.edu}\\ \cline{1-4}
SOFA    & \multicolumn{1}{l|}{SCN}     &                                                      & Real-time soft-body simulation and surgical interaction modeling for indoor scenarios; Url: \url{https://www.sofa-framework.org}\\ \cline{1-4}
Project Chrono & \multicolumn{1}{l|}{JSON}    & URDF                                                 & Simulates rigid/flexible bodies, fluids, and granular materials; advanced multibody physics; Url: \url{https://projectchrono.org}\\ \cline{1-4}
Omniverse & \multicolumn{1}{l|}{USD}     & MJCF URDF & GPU-accelerated and photorealistic rendering; built for robotics digital twins with Isaac Sim; Url: \url{https://www.nvidia.com/en-us/omniverse/}\\ \cline{1-4}
Unreal Engine & \multicolumn{1}{l|}{UASSET}  & USD                                                  & High visual fidelity, immersive VR/AR support, and accurate physics for indoor scene interaction; used for robotics, games, architecture; Url: \url{https://www.unrealengine.com}\\ \cline{1-4}
Unity & \multicolumn{1}{l|}{YAML} & & Fast-prototyping; cross-platform; ML and robotics integration via Unity Robotics Hub; Url: \url{https://unity.com}\\ \cline{1-4}
Godot & \multicolumn{1}{l|}{TSCN} & & Open-source and customizable; good for simple 3D/2D simulations; low resource use; Url: \url{https://godotengine.org}
\\ \cline{1-4}
Blender & \multicolumn{1}{l|}{BLEND} & USD & 
High-quality asset creation and scripting for synthetic data generation in robotics pipelines; large graphics community; Url: \url{https://www.blender.org}
\\ \cline{1-4}
Genesis & \multicolumn{1}{l|}{USD} & URDF MJCF & GPU-accelerated simulation with built-in generative tools for photorealistic indoor robotics datasets; Url: \url{https://genesis-embodied-ai.github.io}\\ \cline{1-4}
\end{tabular}
\label{tab:simulators-comparison}
\end{table*}

\gn{
Physical properties such as mass, friction, stiffness, and viscosity can be derived from videos using simulation and deep learning, as demonstrated in \textit{Scalable real2sim} \cite{pfaff2025scalable}. For rigid bodies, MCMC-based approaches like Galileo \cite{wu2015galileo} estimate mass and friction by matching simulated and observed motion. Hybrid systems \cite{link2022predicting} refine estimates through optimization or neural correction using physics engines.

Soft-body modeling uses learning-based estimations, as in \cite{mao2024video}, which predicts cloth stiffness from video using Transformer architectures and pre-trained networks, achieving 0.995 accuracy. Earlier approaches \cite{bouman2013estimating} used motion cues to infer fabric stiffness and weight via regression.

For fluids, \cite{walker2023go} applies 3D CNNs to predict viscosity from video-recorded flow, outperforming human estimates.

These methods show that deep learning combined with simulation enables automatic, non-invasive inference of physical parameters from visual data across rigid, soft, and fluid domains.
}

\subsection{Introducing Physical Properties in \ac{3dgs} Representations}

\gn{
Recent research integrates physical simulation directly into 3D Gaussian frameworks, eliminating the need for traditional meshes. PhysGaussian \cite{xie2024physgaussian} embeds Newtonian dynamics within 3D Gaussians using a modified Material Point Method (MPM) that introduces deformation and stress attributes governed by continuum mechanics. Both physical simulation and visual rendering share the same Gaussian kernels, ensuring consistent dynamics and appearance.

}

\am{
\subsection{Open Challenges}
Hybrid deep-learning and differentiable-simulation methods now enable automated, non-invasive inference of material and dynamic parameters, bridging perception and physical simulation. Future work should focus on developing \textit{physically executable digital twins} that couple high-fidelity visual models (e.g., 3DGS) with physics-aware layers for realistic contact, deformation, and fluid interactions. Persistent challenges include the absence of unified scene description standards (URDF–USD interoperability), and limited generalization to mixed rigid–soft–fluid domains. Promising directions involve hybrid 3DGS–mesh representations for joint visual and physical reasoning, automated extraction of collision and material properties, and online learning for real-time simulation refinement. Establishing open datasets, standardized scene schemas, and integrated toolchains will be crucial for advancing physically grounded digital twin systems.
}

\section{Semantics}
\label{semantics}

The integration of semantic information into digital twins extends their utility beyond geometric and physical fidelity by enabling reasoning about purpose, functionality, and inter-object relationships. The representations by which semantics are encoded range from symbolic structures such as ontologies and knowledge graphs to deep neural networks, and permit downstream applications such as planners to make inferences about the behaviors, uses and meaning of the represented physical entities.

\subsection{Structural Semantics}
\label{sec:structural-properties}

\subsubsection{Semantic Scene Graphs}
\label{sec:semantic-scene-graphs}

\Acp{ssg} provide structured, relational representations of a scene by encoding object instances as nodes and their spatial, semantic, or functional relationships as edges. They serve as intermediate representations that support reasoning, planning, and interaction, integrating both perceptual and symbolic information.

In the 3D domain, \ac{ssg} construction typically begins from point cloud data or RGB-D sequences. Wald et al.~\cite{wald2020learning} proposed constructing \acp{ssg} from 3D point clouds after performing class-agnostic instance segmentation. In a follow-up work, Wald et al.~\cite{wald2022learning} extended this approach by jointly learning instance embeddings, thus obviating the need for pre-computed segmentation masks. Wu et al.~\cite{wu2021scenegraphfusion} presented SceneGraphFusion, which incrementally builds 3D scene graphs from RGB-D video using PointNet and a graph neural network. Similarly, Wu et al.~\cite{wu2023incremental} demonstrated 3D \ac{ssg} construction from monocular 2D video. Active perception strategies have been employed to optimize \ac{ssg} construction: Li et al.~\cite{li2022embodied} propose to learn exploration policies that guide the agent to acquire views conducive to better \ac{ssg} estimation. InstructScene \cite{lin2024instructscene} uses a text encoder and graph transformer to build \acp{ssg} from high-level spatial relations, followed by a diffusion model to reconstruct the 3D geometry of the scene.

Several systems go beyond pure perception to integrate \acp{ssg} into reasoning and control. EmbodiedVSR~\cite{zhang2025embodiedvsr} fuses dynamic scene graph generation with embodied spatial reasoning and robotic control by constructing an \ac{ssg} that includes objects, attributes, and ``scaleless spatial relations,'' and uses a textual representation of this graph as input to a large language model for downstream tasks. SGFormer~\cite{lv2024sgformer} combines 3D point cloud features with semantic embeddings obtained via ChatGPT, and includes graph embedding and semantic injection layers to produce an \ac{ssg}. Mania et al.~\cite{10610743} developed RoboKudo, which integrates \ac{ssg} construction into the robotic perception-action loop using behavior trees and the \ac{uim} architecture. It is integrated with a \ac{vr} engine for photorealistic and physics-aware simulation~\cite{kumpel2021semantic}. In 3D+time (4D) video settings, Yang et al.~\cite{yang20234d} presented PSG4DFormer, a unified framework that combines 4D panoptic segmentation with relation modeling to construct temporally consistent panoptic scene graphs. This work also introduced the PSG-4D dataset for benchmarking.

Together, these works demonstrate the central role of \acp{ssg} in bridging perceptual input and symbolic reasoning. As structured, interpretable, and extensible representations, \acp{ssg} provide a foundational mechanism for embedding semantic understanding into digital twins, particularly in robotics applications where perceptual grounding and task-driven interaction are critical.

\subsubsection{Knowledge Graphs}
\label{sec:knowledge-graphs}

Semantic scene graphs offer spatially grounded, instance-level representations of observed scenes. \Acp{kg}, by contrast, encode entities and their interrelations at a higher level of semantic abstraction, facilitating reasoning over conceptual and relational knowledge that may not be directly perceptible. In the context of digital twins, \acp{kg} complement scene graphs by enabling symbolic inference, task-level grounding, and integration with external knowledge repositories. We refer to \cite{ji2021survey} for a comprehensive survey on general methods for \ac{kg} construction.

A key capability in bridging visual perception with symbolic representations is \textit{\ac{pel}}: the task of connecting perceptual observations to existing entities in a target \ac{kg}. Adamik et al. \cite{adamik2024advancing} propose a complete \ac{pel} pipeline that performs object detection using YOLO, constructs an intermediate ``observation graph,'' and resolves entity matches by comparing object label similarities and disambiguating candidates using semantic distance in WordNet.

\textit{\ac{mel}} extends \ac{pel} to multiple modalities: Shi et al.~\cite{shi2023generative} propose a generative architecture with a feature mapper that aligns vision-language embeddings with \ac{kg} concepts. Zheng et al.~\cite{zheng2022visual} present a deep ranking model trained to select the \ac{kg} concept most consistent with visual and linguistic features.

Several works explore \textit{neuro-symbolic approaches} that internalize latent symbolic structures. The Neuro-Symbolic Concept Learner (NS-CL)~\cite{mao2018the} acquires object-level scene representations that can be queried in an open-vocabulary fashion without direct supervision on symbolic concepts. ZeroC~\cite{wu2022zeroc} associates visual inputs with a symbolic \ac{kg} through an energy-based model for zero-shot concept acquisition. NeSy-$\pi$~\cite{sha2024neuro} incrementally discovers new relational concepts by learning neural rule and concept modules from 2D images in synthetic tabletop domains. By contrast, Alberts et al.~\cite{alberts2020visualsem} provide pretrained neural models for concept retrieval over a large-scale multimodal \ac{kg} that links images, concepts, and named entities through typed semantic relations.

\subsubsection{Articulation}

\begin{figure*}[t]
\centering
\includegraphics[width=\linewidth]{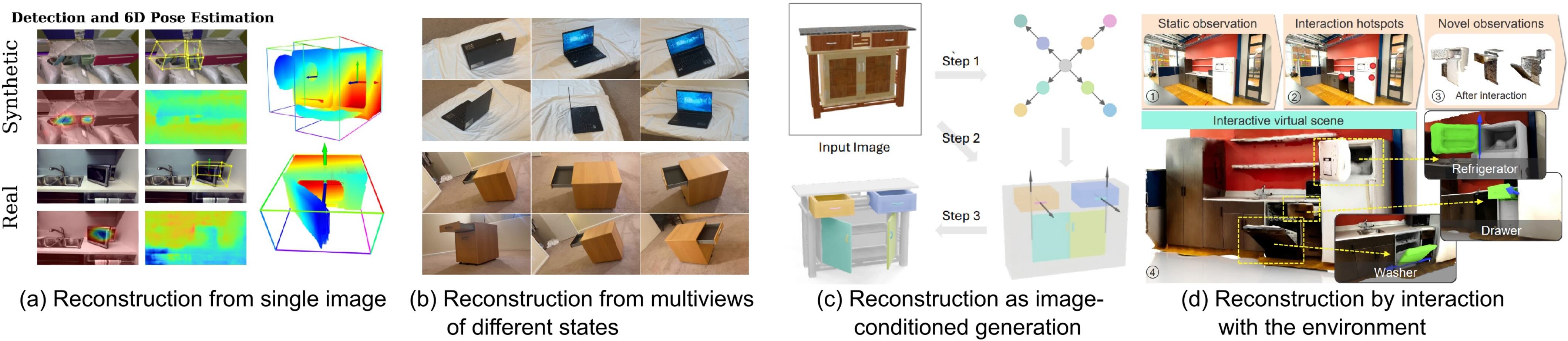}
\caption{Articulated object reconstruction from different visual inputs. Figures reproduced from original papers~\cite{heppert2023carto, wei2022self, liu2024singapo, hsu2023ditto}.
}
\label{fig:articulation}
\end{figure*}

\begin{table*}
\caption{
Datasets related to articulated objects or scenes
}
\vspace{-0.05in}
\centering
{
\begin{tabular}{@{} |l|l|c|c|c|r|r|@{}}
\hline
\textbf{Dataset} & \textbf{Source}    & \textbf{Image}    & \textbf{Texture}    & \textbf{\# classes}   & \textbf{\# objects}    & \textbf{\# parts}   \\ 
\hline
RPM-Net~\cite{yan2020rpm}                   & Synthetic      & \xmark              & \xmark         & 43    & 969           & 1,420          \\
Shape2Motion~\cite{wang2019shape2motion}    & Synthetic      & \xmark              & \xmark        & 45     & 2,440         & 6,762         \\
PartNet-Mobility~\cite{xiang2020sapien}     & Synthetic       & \xmark              & \cmark         & 46    & 2,346         & 14,068        \\ 
AKB-48~\cite{liu2022akb}                 & Realistic        & \xmark              & \cmark         & 48    & 2,037         & -             \\ 
OPDReal~\cite{jiang2022opd}             & Realistic         & \cmark       & \cmark         & 8     & 284           & 875           \\ 
MultiScan~\cite{mao2022multiscan}      & Realistic           & \cmark       & \cmark          & 20    & 10,957        & 5,129        \\ 
GAPartNet~\cite{geng2023gapartnet}       & Synthetic \& Realistic       & \xmark              & \cmark          & 27    & 8,489         & 1,166        \\
ACD~\cite{iliash2024s2o}                & Synthetic      & \xmark              & \cmark          & 21    & 354           & 1,350        \\
\hline 
\end{tabular}
}
\label{tab:articulation_datasets}
\end{table*}

Articulated object reconstruction aims to determine the 3D structure and motion parameters of objects based on diverse observations, such as point clouds, multi-view RGB or RGB-D images, and RGB videos (see Fig. \ref{fig:articulation}). Two main approaches exist: modeling an object as a single deformable surface across time or states~\cite{mu2021sdf, wei2022self, heppert2023carto}, or reconstructing each component as distinct surfaces linked by joints~\cite{tseng2022cla, jiang2022ditto, liu2023paris}. A-SDF~\cite{mu2021sdf} pioneered the use of signed distance fields to reconstruct articulated objects from a single image, while subsequent works expanded the input modality to multi-view images~\cite{wei2022self}, stereo pairs~\cite{heppert2023carto}, or interpolated states without prior knowledge of movable parts~\cite{swaminathan2024leia}. However, these surface-based methods often lack part-specific geometry, complicating their use in simulation. To address this, part-level methods such as CLA-NeRF~\cite{tseng2022cla}, Ditto~\cite{jiang2022ditto} and PARIS~\cite{liu2023paris} reconstruct multiple articulated parts, while Real2Code~\cite{mandi2024real2code} offers more flexibility by focusing on cuboid-like parts without assumptions on their number. Beyond static reconstruction, Hsu et al.~\cite{hsu2023ditto} jointly learn affordance and articulation models through interaction with the environment.

Complementary to reconstruction, articulated object \textit{generation} seeks to synthesize both geometric and kinematic structures. NAP~\cite{lei2023nap} generates full object descriptions including articulation information unconditionally from noise, while CAGE~\cite{liu2024cage} introduces a conditional setting based on category and part connectivity graphs. SINGAPO~\cite{liu2024singapo} focuses on household objects, generating hierarchical articulated models from part-panoptic segmentation or even from a single image. However, its closed vocabulary and limited articulation types constrain generality. Articulate-Anything~\cite{le2024articulate} provides a more general framework for articulating objects from diverse modalities such as text, images, and videos, using programmatic composition to ensure plausible articulation. Table \ref{tab:articulation_datasets} provides an overview of common datasets for training and benchmarking articulation models.

\subsection{Affordances and Dispositions}
\label{sec:affordances}

Affordances describe the possible interactions an agent can perform with an object, and are thus essential for constructing functional digital twins that support task planning, interaction modeling, and simulation: Every physical interaction leverages at least one affordance ~\cite{toumpa2023object}. We refer to Hassanin et al.~\cite{hassanin2021visual} and Chen et al.~\cite{chen2023survey} for comprehensive surveys of visual affordance recognition. Among other challenges, they highlight a need for robust generalization to real-world conditions as well as lack of high-quality supervised training data. Consequently, recent work emphasizes open-vocabulary and zero-shot affordance inference. For example, Shao et al.~\cite{shao2024great} enhance affordance grounding using chain-of-thought reasoning with \acp{vlm} to systematically identify interaction parts, geometric features, and intended uses. Robo-ABC~\cite{ju2024robo} bootstraps affordances for novel objects by analogy with known objects stored in a memory structure. By contrast, Li et al.~\cite{li2024one} leverage large-scale foundation models for open-vocabulary zero-shot generalization. In the 3D domain, Nguyen et al.~\cite{nguyen2023open} propose an open-vocabulary affordance segmentation framework that operates directly on 3D point clouds. Toumpa et al.~\cite{toumpa2023object} use object detection networks and graph embeddings to construct Activity Graphs (AGs) from 3D video, which explicitly represent spatiotemporal relations between objects and permit the inference of object affordances in ways that are robust to scene-specific variations, such as object colors or distances between objects. With particular relevance to digital twins for robotic manipulation, Yu et al.~\cite{yu2024uniaff} unified the prediction of grasp and functional affordances, along with articulation parameters, in a single model.

Collectively, these approaches illustrate a trend towards more generalizable, semantically rich, and multimodal affordance models that are essential for constructing interactive and functional digital twins capable of supporting downstream reasoning and control.

\subsection{Implicit Semantics}
\label{sec:implicit-semantics}

Traditional approaches to semantic scene understanding in digital twins rely on explicit representations such as symbolic labels, scene graphs, or structured affordance models. In contrast, a wide range of recent approaches operate on learned representations, typically encoded in neural networks, that capture semantics without requiring predefined taxonomies, enabling open-vocabulary queries.

\subsubsection{Scene Understanding}
\label{sec:implicit-semantics-scene-understanding}
Foundational \acp{vlm} like CLIP~\cite{radford2021learning}, DINO~\cite{caron2021emerging}, and SAM~\cite{kirillov2023segment} provide a modular basis for class- and part-level semantic grounding. CLIP-GS~\cite{Liao2024} integrates CLIP features with \ac{3dgs} for open-vocabulary class-level understanding. SGS-SLAM~\cite{Li2024f} uses 2D panoptic segmentation to guide \ac{3dgs}, producing semantically segmented 3D reconstructions. Similarly, EgoLifter~\cite{Gu2024} augments \ac{3dgs} with semantic features learned from SAM. 

Several works combine \ac{3dgs} with open-vocabulary features from CLIP, SAM, and DINO to construct \textit{semantic feature fields} or \textit{feature-carrying Gaussians}, as in SemGauss-SLAM~\cite{Zhu2024} and SAGD~\cite{Hu2024}. These fields support spatially grounded semantic queries, semantic \ac{slam}, and improved 3D boundary segmentation. Distilled Feature Fields (DFFs)~\cite{kobayashi2022decomposing} adapt these ideas to NeRF representations, while Feature Splatting~\cite{Qiu2024} extends semantic feature fields to encode object dynamics and physics parameters. The Large Spatial Model (LSM)~\cite{Fan2024b} integrates 2D vision and language inputs to directly predict semantic Gaussians from unposed images.

Cross-modal embedding approaches have scaled implicit semantic reasoning to larger datasets. ULIP~\cite{xue2022ulip,xue2023ulip2}, OpenShape~\cite{liu2023openshape}, and Uni3D~\cite{zhou2023uni3d} align 2D images, 3D point clouds, and open-vocabulary text in shared feature spaces, enabling applications such as text-based retrieval, editing, and segmentation. Extensions include multi-view training~\cite{lee2024duoduo}, contrastive text-to-shape retrieval~\cite{ruan2024tricolo}, and cross-modal alignment via Earth Mover’s Distance~\cite{ren2025sca3d}.

\subsubsection{Scene Generation}
\label{sec:implicit-semantics-scene-generation}
Beyond scene understanding, implicit semantics are also used for scene generation. CAST~\cite{yao2025cast} uses GPT-based models to generate physically coherent object layouts from open-vocabulary prompts, incorporating physics-aware corrections based on commonsense spatial relations. SceneEval~\cite{tam2025sceneeval} proposes a framework for evaluating scene coherence with requirement-specific reasoners, quantifying both explicit and implicit goal fulfillment.

In related work, Dai et al.~\cite{dai2024automated} introduce ACDC, a pipeline for constructing Digital Cousins from a single image, using DINO and GPT to segment, label, and orient matched assets from a database. These assets include articulated components with known functional dispositions (e.g., drawers that can be opened). Mu et al.~\cite{mu2025robotwin} adopt a similar paradigm for constructing semantically annotated digital twins of individual objects from single RGB images, comprising 3D geometry and textures, using generative models. Their model automatically annotates function axes, contact points, and manipulation affordances. An \ac{llm} then uses this structured annotation to synthesize manipulation trajectories for downstream robotic execution.

\subsection{Open Challenges}

Recent progress on open-vocabulary panoptic segmentation with \acp{vlm} have enabled rapid progress on digital twin technologies for modeling of scene, object and agent semantics. Due to the limitations of foundational technologies as well as the complexity of real-world use cases, several open questions remain. Probably the most important current limitation is \textit{robust generalization}. The current generation of foundation models is not robust to out-of-distribution (OOD) data and prone to hallucinations \cite{rawte2025defining}, leading to incorrect or unstable scene descriptions in situations that are underrepresented in their training dataset. Methods to reliably extract semantics under data scarcity as well as approaches for validating the correctness or plausibility of extracted semantics are important avenues for future work. Other important avenues for future work are methods to resolve ambiguous semantics based on context, as well as the social aspects of semantics such as human preferences as well as social norms and expectations.

\section{Conclusion}
\label{sec:conclusion}

While the potential of digital twins has long been established in a wide range of application domains, their usefulness has been limited by the need for prior modeling. Advances in 3D and 4D sensor technologies, generative modeling and multimodal learning now enable the creation of increasingly accurate, and increasingly useful, digital twins purely from visual input. This survey has outlined the evolution of these methods. It focused on 3D Gaussian Splatting as a unifying representation that bridges geometric accuracy, photorealism, and real-time performance. By integrating sparse-view reconstruction, dynamic scene modeling, physical reasoning, and semantic understanding, current research is transforming digital twin creation from a specialized reconstruction problem into an increasingly multimodal, end-to-end process.

One major macro-trend is a convergence toward hybrid frameworks that combine explicit geometry or semantics with neural representations, leveraging large-scale foundation models for perception, language grounding, and physics inference. At the same time, increasingly capable generative models enable digital twin generation from consumer-grade video data, reducing dependence on specialized sensors and manual modeling. However, persistent challenges remain: (1) the integration of lighting, material, and dynamic properties in physically consistent rendering; (2) conversion between learned and standard representations such as USD; (3) reliable temporal modeling and semantic consistency under sparse and noisy observations; and (4) robust generalization to complex, unstructured environments.

\bibliographystyle{IEEEtran}
\bibliography{IEEE}

\end{document}